\documentclass{article} 
\usepackage{colm2024_conference}

\usepackage{microtype}
\usepackage{hyperref}
\usepackage{url}
\usepackage{booktabs}

\title{
Using LLMs to Model the Beliefs and Preferences of Targeted Populations
}
\colmfinalcopy

\author{
\parbox{\linewidth}{
Keiichi Namikoshi \&
Alex Filipowicz \&
David A. Shamma \&
Rumen Iliev \\ \&
Candice L. Hogan \&
Nikos Ar\'echiga
}\\
Toyota Research Institute\\
\texttt{\{keiichi.namikoshi, alex.filipowicz, ayman.shamma, rumen.iliev, }\\\texttt{\quad candice.hogan, nikos.arechiga\}@tri.global} \\
}

\usepackage{graphicx}
\usepackage{amsmath}
\usepackage{amsthm}
\usepackage{bm}
\usepackage{ascmac}
\usepackage{subcaption}
\usepackage{color}

\newcommand{\tref}[1]{Table~\ref{#1}}
\newcommand{\fref}[1]{Figure~\ref{#1}}
\newcommand{\eref}[1]{Equation~\ref{#1}}
\newcommand{\aref}[1]{Appendix~\ref{#1}}
\newcommand{\vect}[1]{\boldsymbol{#1}}
\newcommand{\figcaption}[1]{\def\@captype{figure}\caption{#1}}
\newcommand{\tblcaption}[1]{\def\@captype{table}\caption{#1}}

\usepackage{scalerel}
\def\battery{\scalerel*{\includegraphics{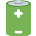}}{\textrm{\textbigcircle}}}
\def\unamusedface{\scalerel*{\includegraphics{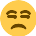}}{\textrm{\textbigcircle}}}
\def\car{\scalerel*{\includegraphics{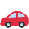}}{\textrm{\textbigcircle}}}

\begin{document}

\maketitle

\begin{abstract}
We consider the problem of aligning a large language model (LLM) to model the preferences of a human population. Modeling the beliefs, preferences, and behaviors of a specific population can be useful for a variety of different applications, such as conducting simulated focus groups for new products, conducting virtual surveys, and testing behavioral interventions, especially for interventions that are expensive, impractical, or unethical.
Existing work has had mixed success using LLMs to accurately model human behavior in different contexts.
We benchmark and evaluate two well-known fine-tuning approaches and evaluate the resulting populations on their ability to match the preferences of real human respondents on a survey of preferences for battery electric vehicles (BEVs).  We evaluate our models against their ability to match population-wide statistics as well as their ability to match individual responses, and we investigate the role of temperature in controlling the trade-offs between these two. Additionally, we propose and evaluate a novel loss term to improve model performance on responses that require a numeric response.
\end{abstract}

\section{Introduction}
In the last decade, large language models (LLMs) have evolved significantly for natural language processing tasks, code generation, and as conversational UIs~\citep{zhao2023survey}.
Existing work generally finds that it is possible to elicit strong agreement between LLM responses and human responses~\citep{dubois2023alpacafarm}. 
This concordance suggests a possibility to
use LLMs as a statistical proxy to study human beliefs, preferences, and behaviors.
This would allow leveraging models that
were trained on large, internet-scale datasets in narrow domains where comparatively small amounts of data are
available.
This might be used, for example, for a company to leverage a comparatively small survey sample
to understand customer preferences with respect to possible new products, to conduct virtual surveys, and to pilot behavioral interventions, such interventions to drive the adoption of 
sustainable technology. The value for intervention research is especially notable in the case when interventions would be impractical or unethical, such as building large amounts of infrastructure or restricting access to infrastructure. To enable such applications, it is crucial to ensure
that LLMs exhibit behavior that is a statistically accurate model of real human behaviors.

Existing literature on this matter finds conflicting results. For example,~\citet{serapiogarcía2023personality} find that it is possible to prompt LLMs in the pathways language model (PaLM) family to exhibit consistent and clearly measurable personality traits. 
Conversely,~\citet{Gui_2023} find significant challenges in emulating human behaviors, specifically in the context of simulated demand for Coca-Cola. These experiments show that 
in general, it is difficult to say \emph{a priori} whether a pre-trained LLM will accurately model a behavior of interest. In this paper, we consider the problem of aligning the beliefs and preferences of a language model so that it can serve as a statistical proxy for a real human population. We consider a macro, population-wide metric as well as a micro, per-individual metric, and we propose a novel penalty term in the loss function to improve performance on numerical survey questions.

We emphasize that the ultimate goal of this work is \emph{not} to produce a survey-answering bot, but instead to align a language model with the beliefs and preferences of real humans as expressed in a survey. The ultimate goal is to arrive at interactive models that enable the study of a target population. Our results indicate that it is easier to model population-wide statistics than individuals, suggesting that one-on-one interviews may be difficult to replicate. However, our population-wide models may still be useful in the context of population-wide studies, for example in the context of marketing, or community-wide simulations, such as those of ~\cite{park2023generative}.

In our experiments, we leverage an existing survey on human beliefs and preferences about battery-electric vehicles (BEVs)~\citep{arechiga2022understanding}. This survey includes interventions intended to increase the preference for BEVs. The contributions of this paper are as follows.
\begin{itemize}
    \item Using the survey data of ~\cite{arechiga2022understanding} on real human study participants, we demonstrate the use of parameter-efficient fine-tuning techniques to improve the agreement of LLMs to human preferences as expressed in survey data.
    \item We investigate the effects of model size, and find that larger pre-trained models provide the best out-of-the-box performance, but this advantage largely disappears after fine-tuning.
    \item We investigate the effects of quantization and sampling temperature. We find that quantizing fine-tuning techniques such as QLoRA~\citep{qlora} provide minimal degradation but large savings in computation. We find that the temperature parameter allows trading off between agreement with population-wide statistics vs matching per-individual responses.
    \item We propose and evaluate a novel penalty term on the loss function to improve model performance on survey questions that require a numerical response.
\end{itemize}
We benchmark against two baseline algorithms trained \emph{de novo} on given survey data, and demonstrate that the fine-tuned LLMs are able to outperform these baseline algorithms under specific settings. Section \ref{sec:related_work} describes related work, Section \ref{sec:background} provides the technical background of our work. Section \ref{sec:problem} describes the problem statement and Section \ref{sec:method} describes our proposed approach. Section \ref{sec:experiments} presents our experiments and Section \ref{sec:conclusions_futurework} concludes and describes directions for future work.

\section{Related Work}\label{sec:related_work}
The possibilities of simulating humans and human behaviors using LLMs~\citep{kaddour2023challenges}, or role-play~\citep{shanahan2023roleplay,wu2023large}, have been discussed in recent work. By applying established psychometrics, 
\citet{serapiogarcía2023personality} demonstrated that LLMs can reliably simulate personalities and that LLM-generated personality traits can be shaped and controlled to imitate specific personality profiles.
\citet{dillion2023can} showed a strong alignment between GPT-3.5 and humans in moral judgments, with a correlation of 0.95. In addition, LLM-based generative agents, when organized as a collective in an interactive sandbox environment, were found to be able to produce believable behaviors not only on an individual level but also on a social level~\citep{park2023generative}.

The ability of LLMs to generate human-like personalities, judgments, and behaviors hints at the opportunity of constructing synthetic human participants in behavioral studies. Several recent works show initial attempts in this direction.
\citet{aher2023using} applied LLMs to simulate human subjects and found that they can reproduce three out of four economic, psycholinguistic, and social psychology experiments and replicate findings from prior studies with real human participants.
\citet{hamalainen2023evaluating} evaluated LLMs' potential of generating synthetic human-computer interaction research data in the form of open-ended questionnaire responses and revealed their capability of generating plausible, human-like self-report data regarding subjective experiences. 

However, previous work simply measured concordance between LLMs and participant data, and other work such as that of ~\citet{Gui_2023} find a lack of agreement in domains such as predicting product pricing. Our work provides a framework to align LLMs to human preferences and explores various techniques to improve the level of agreement.

\section{Background}\label{sec:background}
\noindent\textbf{Auto-regressive large language models}

Autoregressive language models learn to predict the next token in a stream of language tokens. Formally, given a sequence of tokens
$y_1, y_2, \dots, y_n$, the model learns to predict a probability distribution $p(y_{n+1}| y_1, y_2, \dots y_n)$. An important property of these models is
that they can be trained in an unsupervised way, i.e., no manually produced labels are required. All that is required is a large corpus of natural text. Many such
corpora have been assembled from the internet, for example~\citet{pile} and~\citet{together2023redpajama}. A commonly used architecture is the
transformer architecture~\citep{vaswani2017attention}, and specifically the decoder-only transformer with a causal mask, which prevents the model from using information from future tokens~\citep{gpt1}.

\noindent\textbf{Fine-tuning large language models}

A common approach to use LLMs is to pre-train on a large text corpus, and then fine-tune the resulting model for a specific downstream task~\citep{gpt1,bert}. Downstream tasks may include sentiment analysis, question answering, text summarization, etc. The fine-tuning procedure involves updating all of the parameters of the model, and can be computationally expensive for large model sizes.

Low-Rank Adaptation (LoRA)~\citep{lora} is a technique for fine-tuning large language models
that relies on freezing pre-trained model weights and adding low-rank trainable matrices
at various points throughout the model. This procedure dramatically reduces the 
computational cost of fine-tuning since only the low-rank adaptation matrices
have associated gradients at training time.

Quantized LoRA (QLoRA)~\citep{qlora} is a variation of LoRA that quantizes the model weights
as 4-bit NormalFloats, a data type that efficiently compresses the model weights
while discarding as little information as possible. QLoRA also introduces a number of memory optimization techniques, such as double quantization and paged optimizers to manage memory spikes.


\section{Problem setting} \label{sec:problem}
The problem we consider assumes that a small amount of survey data is available from
a representative sample of a target human population. We further assume that
demographic information that is relevant to characterizing the target population is available. 
Formally, the answer $a_{ij} \in \mathcal{A}$ of a participant $i$ who has demographics $x^{\rm d}_i$ is generated with $a \sim p(a | x^{\rm d}_i, x^{\rm q}_j)$.
$x^{\rm q}_j$ denotes the questionnaire $j$.
Demographics, such as age, gender, income, etc., are characteristics of each individual participant. The available demographics as well as their distribution within the target
data should be appropriate to the modeling task at hand.

As a specific example of the survey, we will use the EV-shift dataset~\citep{arechiga2022understanding}.
The EV-shift dataset examines the impact of interventions on the preference for electric vehicles (EVs) when compared to internal combustion vehicles.
This dataset resulted from a study aimed at identifying how effectively different text-based interventions changed people's preferences for EVs.
\tref{tab:evshift_dataset} shows the number of answers and tokens in this dataset.
In the study, subjects began by providing an \emph{initial preference} for EVs, which was a numerical rating from 0 to 100 (with higher numbers indicating greater preference for EVs). Subjects were then shown one of 35 text-based interventions aimed at increasing their preferences for EVs. After the intervention, subjects provided a \emph{post-intervention preference}, which was also a numerical rating from 0 to 100.

Each subject also provided demographic information. 
In total, the dataset contains demographic information, one initial preference rating for each of the 4,045 subjects, the interventions seen by each subject (5 for most subjects), and the post-intervention preference ratings provided by subjects after each intervention.

\begin{table}
\centering
\begin{tabular}{lrr}
\toprule
Questionnaire & \#answer & \#token [B] \\
\midrule
Initial preference & 4,045 & 0.87 \\
Post-intervention preference & 20,217 & 4.79 \\
\bottomrule
\end{tabular}
\caption{Number of answers and tokens in EV-shift dataset. The number of tokens is the amount of tokenized prompt texts that is calculated by the tokenizer of Llama 2~\citep{touvron2023llama}. Procedure to convert survey data to prompt described at section \ref{sec:method}.}
\label{tab:evshift_dataset}
\end{table}

\section{Proposed method}\label{sec:method}
Our proposed approach provides virtual survey participants with a prompted, fine-tuned LLM
to generate survey responses that statistically match those of a target population.
In this section, we describe the implementation of the subjects by prompting and the fine-tuning procedure.

\noindent\textbf{Formalization and implementation} \label{subsec:implementationOfVP}

Each virtual participant in the population is implemented by prompting an LLM to behave as a person with given demographic information.
Formally, the virtual participants with a language model represent the distribution $p(\vect{y} | \vect{x})$, where $\vect{y} \in \mathcal{V}^N$ is the generated token sequence. This generated token sequence corresponding to the answer from token sequence $\vect{x} \in \mathcal{V}^M$ corresponding to the demographics and survey question.
$\mathcal{V}$ is the vocabulary of the language model.
Additionally, let $F$ be a function that preprocesses the output sequence $\vect{y}$ to produce an answer $a$. This function is useful when the expected answer to a survey question is a structured output (e.g., a numerical rating, or a multiple choice answer), but the LLM embeds its answer within a longer explanation. The complexity of $F$ may become quite high. In our experiments, we adhere to a simple function that allows some flexibility interpreting the model outputs, without becoming overly complex. We define 
$F: \mathcal{V}^N \to \mathcal{A}$ as the function that extracts the first (possibly multi-digit) number that appears in the symbol sequence as a simple implementation.


\begin{figure}[htbp]
\centering
\includegraphics[width=1.0\columnwidth]{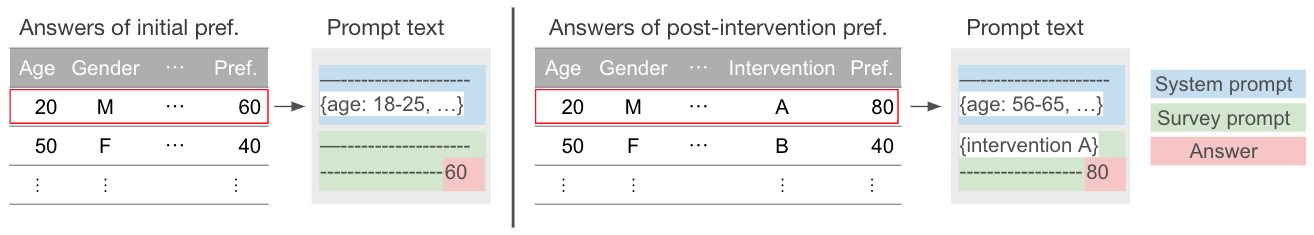}
\caption{Convert survey data to prompt text. Left: Initial preference questionnaire. Right: Post-intervention preference questionnaire.}
\label{fig:proposed_method}
\end{figure}

\fref{fig:proposed_method} shows an example of formulating the prompt text from survey data.
To set the properties of the virtual participant, we use the following prompt,
\begin{itembox}[l]{System prompt}
I want you to act as the following character. Answer all of the following questions from the point of view of this character, do not break character. \{demographics\}
\end{itembox}
\{demographics\} contains a list of demographic characteristics. 
For example, a man aged 18--25 is represented as \{age: 18--25, gender: man\}.
Our implementation allows generating virtual participants with demographics drawn from any distribution, enabling targeting to different populations of interest, including demographics that a company believes are likely to be customers for a specific product. Although prompt engineering techniques are a rich area of inquiry, they are not the focus of our work, and for this reason we limit our prompts to the minimum amount of information required to explain the setting to the LLM. Naturally, our techniques can be combined with prompt engineering techniques to enhance their effectiveness.

Our survey concerns preferences for battery electric vehicles. To check the preferences of a virtual participant given demographics, we use the following prompt,
\begin{itembox}[l]{Survey prompt}
\{intervention\} On a scale from 0 to 100, what is your current preference for battery electric vehicles (BEVs)? Please reply with just a single number rating and no additional words or explanations. Score: 
\end{itembox}
\{intervention\} is blank for the initial preference question, and contains intervention text to produce post-intervention preferences.
For example, intervention sentence is "Some BEV manufacturers may start offering free charging" (interventions are described in Appendix \ref{appendix:setting}).


\noindent\textbf{Fine-tuning large language model with survey data}

Next, the pre-trained LLM is fine-tuned to emulate the preferences of the human survey participants.
The LLMs used for fine-tuning are auto-regressive LLMs for text generation.
Each text corresponds to a set of question and answer for a specific subject, and the text contains a system prompt, a survey prompt, and an answer.
Note that a single dataset contains multiple questions, i.e., both initial preference and post-intervention preference questions.
For most of our experiments, we use the conventional cross-entropy loss function.

\noindent\textbf{Numeric penalty function} 

In \ref{subsec:penalty}, we seek to enhance model performance by adding a penalty term to the cross-entropy loss function. This specific penalty term is novel (to the best of our knowledge), and is represented by \eref{eq:penalty_loss}.

For this penalty term to be well-defined, we require that the vocabulary contain separate tokens for each of the possible numerical output tokens. Since we are using the Llama 2 family of models, and these models have separate tokens for the digits 0 through 9, we need to scale the survey data so that the possible answers are in the single-digit range in order to be able to use this penalty term. 

Our numerical penalty term is calculated by weighting the log generation probability of the answer set containing the subject's answers by a value $w_{a,\hat{a}}$ for each answer. This value is equal to $1$ when the generated numerical answer $a$ is exactly equal to the true answer $\hat{a}$, zero when $a$ is further than a hyperparameter value $d$ from $\hat{a}$, and a computed intermediate value in between, as shown in \eref{eq:penalty_weight}. This hyperparameter controls how much information the penalty term provides to nearby numerical values.

The goal of this penalty term is to improve the performance of a token generation model over questions that require a numerical response. The cross-entropy loss term merely provides feedback about whether a generated numerical token was correct or not. Our penalty term additionally provides feedback about whether the generated numerical token was close to or distant from the correct answer.

The combined loss function is $L(D; \theta) = (1 - \alpha) L^{\rm cross} (D; \theta ) + \alpha L^{\rm penalty} (D; \theta )$. 
$D$ denotes the data, $\theta$ denotes the parameters of the language model, and $\alpha \in [0,1]$ denotes a mixing coefficient between the two loss terms. 


\begin{equation} \label{eq:penalty_loss}
L^{\rm penalty} (D; \theta ) = - \frac{1}{|D|} \sum_{
\substack{\vect{x}, \hat{y} \in D \\ \hat{a} = F(\hat{y})}
} \sum_{a \in \mathcal{A}} 
\frac{w_{a, \hat{a}}}{\sum_{a' \in \mathcal{A}} w_{a', \hat{a}}} \log{p(\hat{y} | \vect{x})}
\end{equation}

\begin{equation}\label{eq:penalty_weight}
w_{a, \hat{a}} = \left\{\begin{array}{ll}
    1 - \frac{1}{d} |a - \hat{a}| & \text{if}~|a - \hat{a}| < d \\
    0 & \text{otherwise}
\end{array}\right.
\end{equation}

\noindent\textbf{Performance Measures}

Finally, we measure the agreement between LLMs and humans.
We use the test data portion of the survey data for this measurement.
The LLM responses are generated using the same demographics distribution as the survey data to be measured. The metrics we use to measure similarity between LLM responses and survey responses are KL-divergence and root mean square error (RMSE).
Intuitively, we can think of the KL-divergence as measuring model agreement with the statistics of the population as a whole, whereas the RMSE measures model agreement with individual responses. 


\section{Experiments}\label{sec:experiments}

In these experiments, we use Llama 2~\citep{touvron2023llama}.
The model sizes are 7B, 13B, and 70B, using chat models published on HuggingFace\footnote{\url{https://huggingface.co/meta-llama}}.
We set LoRA $r=8, \alpha=32$, and LoRA dropout $0.1$ for fine-tuning.
We use 3 epochs across all experiments.
We split our dataset randomly into a training set, a validation set, and a test set using an 8:1:1 split by subject (so the number of subjects in train data is 3,237). Critically, subjects that occur in the training set do not appear in the test set.

Evaluation on the test data was performed using the following procedure. Although the prompt asks the model for a numerical answer, sometimes the model replies with additional text. Since we do not want to construct arbitrarily complex logic to understand all possible
generations, we sample a maximum of 8 output tokens,
and the first number that appears from the beginning of the sentence is considered the answer.
If a number is not included in the generated sentence or is not an integer in the correct range, the generation is considered to have failed.
KL-divergence and RMSE are calculated except in cases where the generation fails. KL-divergence is obtained by generated numbers that are discretized.
The discretization width is 10 in Section \ref{subsec:model_size} to \ref{subsec:temperature} and 1 in Section \ref{subsec:penalty} for adapting a scale of the numerical answer.

To understand the performance of our language models, we compare the model against
three baselines that are not language models.
The first two are supervised learning algorithms, support vector regression (SVR) and CatBoost ~\citep{NEURIPS2018_14491b75}. These algorithms are trained on the survey data, and they learn to map a vector of demographic information to a predicted preference value. The purpose of these benchmarks is to situate the performance of the language models with respect to highly effective supervised learning models. We note that in many configurations cases, the baseline models outperform the language models. For our use-case, however, the supervised learning algorithms cannot be used in downstream tasks, such as follow-up questions that involve conversational responses or user surveys.
The third benchmark model is a model that generates a random answer.
These baselines will be conducted to evaluate the positioning of LLM performance at individual-level and population-level by showing the curve of possible solutions that can be achieved by non-language methods, directly mapping from demographic characteristics to survey responses.
For SVR and CatBoost, we will refer to the resulting curve of performance values from models trained with multiple hyperparameters as the baseline curves.
We chose SVR as one of our baselines because it is a commonly used supervised learning method for regression problems, and CatBoost because it is a powerful gradient-boosting method for categorical variables.
SVR and CatBoost are fitted with the EV-shift dataset for each questionnaire.
For SVR, categorical variables such as demographics and intervention text are converted to dummy variables.
The predicted preference is normalized between 0 to 1 for SVR and CatBoost.
The hyperparameters for SVR and CatBoost are shown in \aref{appendix:baselines}.

We detail our experiments in the sections below Section \ref{subsec:model_size} investigate the effects of model size, Section \ref{subsec:quantization} investigates the effect of model quantization, Section \ref{subsec:temperature} investigates the effects of sampling temperature, and Section \ref{subsec:penalty} investigates the effect of our proposed penalty term.


\subsection{Effects of model size}\label{subsec:model_size}
\begin{figure}[htbp]
\centering
\includegraphics[keepaspectratio, width=1.0\columnwidth]{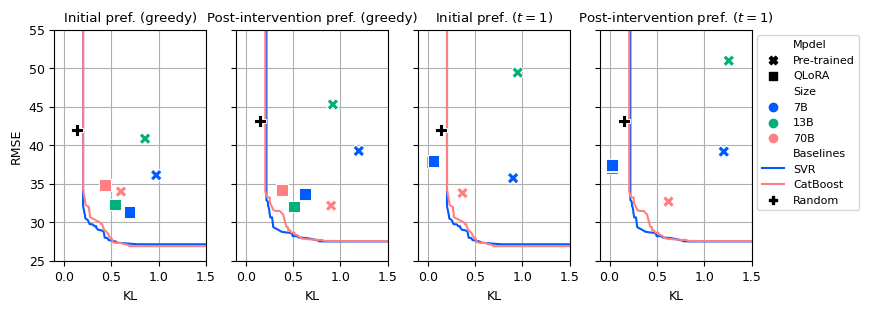}
\caption{Benchmark results. Line plots indicate the baseline points for each hyper-parameter. The left two plots show greedy sampling results ($t=0$). With greedy sampling, the fine-tuned models all outperform the pre-trained models. The largest model attains the best KL-divergence, but not the best RMSE score. None of the models outperform the supervised learning baselines on either RMSE or KL-divergence. The right two plots show calibrated sampling ($t=1$). All models outperform the baselines on KL-divergence, but not on RMSE. The square boxes overlap because the difference in fine-tuned model performance is small. Each value of fine-tuned models is described on \tref{table:prformance_qlora_t1}.
}
\label{fig:kl_rmse_size}
\end{figure}

First, we explore the effects of fine-tuning different model sizes. For this experiment, we will fine-tune three different model sizes (7B, 13B, and 70B) with QLoRA. We trained the model for 3 epochs, but rolled back to the 1 epoch checkpoint due to increases in the validation loss (see Appendix \ref{appendix:modelsize}).
The RMSE-KL plots for each model and baseline are shown in \fref{fig:kl_rmse_size}.
We consider two sampling temperatures at the output of the model, corresponding to a temperature of zero and a temperature of one. These values are chosen due to their natural interpretations. A temperature of zero corresponds to greedily taking the token with the highest output probability, which can be interpreted as the token that the model most strongly believes is the correct value. We denote this setting as \emph{greedy} sampling. A temperature of one corresponds to sampling output tokens with the probability distribution that is obtained by directly applying a softmax function to the logits of the neural network. Since in this case output tokens will appear with a distribution that corresponds exactly to the softmax distribution at the model output, we denote this setting as \emph{calibrated} sampling. We will investigate the role of temperature in greater detail in Section \ref{subsec:temperature}.

Comparing the results of the pre-trained model and QLoRA, the use of greedy sampling tends to reduce both KL-divergence and RMSE, while the use of calibrated sampling significantly reduces the KL-divergence.
Comparing results by model size, 70B had the best (lowest) performance on both metrics among the pre-trained models, and QLoRA improved KL-divergence for all model sizes.
However, when calibrated sampling was used, the QLoRA model outperformed the baseline KL-divergence for all sizes.
In other words, with and without fine-tuning, 70B has a smaller KL-divergence than the other sizes, but the difference is smaller when fine-tuning is used.

These results indicate that fine-tuning not only reduces both RMSE and KL-divergence, but also can exceed the KL-divergence of the non-language model under some sampling conditions. Also, larger models tend to display lower KL-divergence. Although our language models with greedy sampling do not outperform the supervised learning benchmarks, using a language model is more versatile than a supervised learning model, since the language model can be queried with natural-language follow-up questions, or asked to provide natural-language responses in a user interview.

\begin{table}[h]
\caption{Performance of QLoRA with calibrated sampling ($t=1$). Bold values represent the minimum value between model sizes.}
\label{table:prformance_qlora_t1}
\begin{minipage}[t]{.45\textwidth}
\centering
\caption{Initial preference}
\label{table:}
\begin{tabular}{rrr}
\toprule
Size & KL-divergence & RMSE \\
\midrule
7B & 0.063 & 37.914 \\
13B & 0.046 & 37.976 \\
70B & \textbf{0.042} & \textbf{37.799} \\
\bottomrule
\end{tabular}
\end{minipage}
\begin{minipage}[t]{.45\textwidth}
\centering
\caption{Post-intervention preference}
\label{table:}
\begin{tabular}{rrr}
\toprule
Size & KL-divergence & RMSE \\
\midrule
7B & 0.026 & 37.411 \\
13B & 0.025 & \textbf{36.970} \\
70B & \textbf{0.019} & 37.266 \\
\bottomrule
\end{tabular}
\end{minipage}
\end{table}

\subsection{Quantization effects} \label{subsec:quantization}

Next, we show the impact of the choice of fine-tuning method, specifically comparing LoRA~\citep{lora} and QLoRA~\citep{qlora}, which differ mainly in that QLoRA introduces parameter quantization.
In this experiment, we focus on the 7B parameter model.

The comparison of KL-divergence and RMSE for each question is shown in \tref{table:comparison_qlora_lora}.
LoRA tended to produce lower (better) KL-divergence than QLoRA for initial preference questions, but higher (worse) KL-divergence for post-intervention questions. On the RMSE metric, LoRA performs slightly worse than QLoRA on the initial preference questions but slightly better on the post-preference questions.
However, these differences are fairly small. When the preferences for each question were compared, Spearman's correlation coefficients were 0.9 and 0.81, indicating very high correlations.

These results indicate that the effect of quantization on the responses is small. Since QLoRA provides higher computational efficiency at fine-tuning time, our experiments corroborate the view that QLoRA is able to efficiently provide fine-tuning capabilities.

\begin{table}[htbp]
\caption{Comparison between QLoRA and LoRA. (7B, greedy sampling)}
\label{table:comparison_qlora_lora}
\begin{tabular}{lrrrr}
\toprule
 & \multicolumn{2}{c}{KL-divergence} & \multicolumn{2}{c}{RMSE} \\
Method & Initial preference & Post preference & Initial preference & Post preference \\
\midrule
QLoRA & 0.694 & \textbf{0.628} & \textbf{31.308} & 33.703 \\
LoRA & \textbf{0.670} & 0.673 & 31.641 & \textbf{33.091} \\
\bottomrule
\end{tabular}
\end{table}

\subsection{Temperature effects} \label{subsec:temperature}

In this section, we show the impact of the decoding temperature on our performance metrics.
According to \citet{wiher2022decoding}, there are various decoding strategies. In this paper, (ancestral) sampling is used as calibrated sampling.
In these experiments, we fix our attention on a 7B model fine-tuned with QLoRA.

\begin{figure}[htbp]

\begin{minipage}[t]{0.47\linewidth}
\centering
\includegraphics[width=1.0\textwidth]{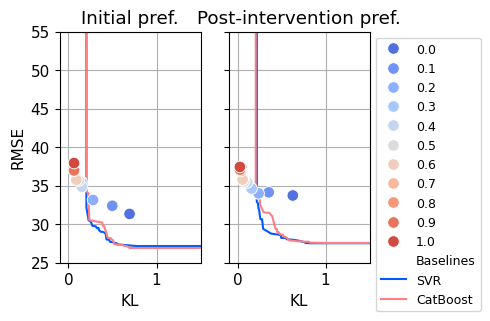}
\caption{Sampling temperature effects. 7B+QLoRA. A temperature of 0.0 corresponds to greedy sampling, and a temperature of 1.0 corresponds to calibrated sampling. Varying the temperature allows trading off the population-wide statistical metric of KL-divergence against the per-individual RMSE metric.}
\label{fig:kl_rmse_temperature}
\end{minipage} 
\hspace{0.02\columnwidth}
\begin{minipage}[t]{0.47\linewidth}
\centering
\includegraphics[width=1.05\textwidth]{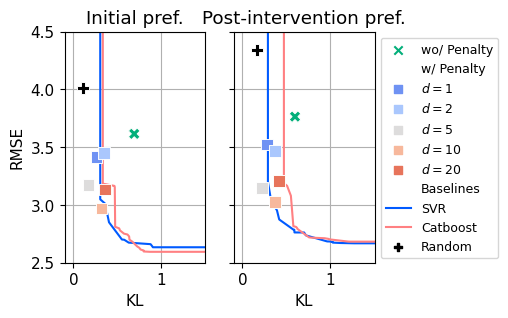}
\caption{Numerical penalty term effects. 7B+QLoRA. The coefficient $\alpha$ of the penalty term is fixed at 0.5.  Penalty term allowed to decrease RMSE, It tends to decrease RMSE the most when $d=10$.}
\label{fig:kl_rmse_slope_width}
\end{minipage}

\end{figure}

The RMSE-KL plots for varying the temperature parameter of the stochastic sampling are shown in \fref{fig:kl_rmse_temperature}.
The results show that greedy sampling has the lowest RMSE, and increasing temperature (and randomness) tends to decrease (improve) KL-divergence and increase (worsen) RMSE.

From these results, we can see that the population-wide metric of KL-divergence and the per-individual metric of RMSE trade off against each other, and that the choice of temperature allows fine-grained control over this trade-off.


\subsection{Numerical penalty term effects} \label{subsec:penalty}

In this section, we show the effect of the penalty term.
The experiment uses a 7B model fine-tuned with QLoRA. We ran the fine-tuning for 3 epochs. The validation loss started to increase after epoch 1 (as detailed in Appendix \ref{appendix:penalty}), so we used the checkpoint at epoch 1. In this experiment, we scaled the preference values to the range 0 to 9. This scaling is performed because the penalty term can only be applied to responses where a numerical response is given by a single output token of the language model, and the Llama 2 tokenizer has distinct tokens for only the digits 0 to 9.

The comparison between the case without penalty term and the case where the hyperparameter $d$ of the penalty term is varied is shown in \fref{fig:kl_rmse_slope_width}.
The results show a tendency to decrease both measures when a penalty term is used compared to the case without a penalty term.
In addition, increasing $d$ tends to decrease RMSE. However, the best value of $d$ is found to be $10$. We fixed the value of $\alpha=0.5$ for this analysis because this produced the best performance in our experiments. Note that $\alpha=0.5$ provides equal weighting to the cross-entropy loss and the numerical penalty term.


\section{Conclusions and Future Work}\label{sec:conclusions_futurework}
We have investigated the use of LLMs to model the beliefs and preferences of a human population. This can be useful, for example, to conduct simulated focus groups for new products, conduct virtual surveys, or pilot interventions that would be unethical or impractical to conduct on real humans. We found that out-of-the box pre-trained models provide comparatively poor performance at predicting the responses of human survey participants, but that the LLMs can be fine-tuned to provide a better model of the target population. We investigated the effects of model size, and found that larger models provide better performance, but this advantage shrinks after task-specific fine-tuning. We investigated the effects of quantization on the fine-tuning process, and found that the resulting degradation was minimal, confirming that quantization is a viable technique to reduce computation costs. We investigated the role of sampling temperature, and found that the temperature allows trading off the population-wide metric of KL-divergence against the per-individual metric of RMSE. Finally, we introduced a penalty loss term to improve the performance of the model on questions that require a numerical output, providing the model with additional information at training time about the relative correctness of different numerical responses.

Although our approach demonstrates that it is possible to match responses on survey data, in future work we will study the extent to which this shift successfully aligns the model to unseen behavioral scenarios. 

\bibliography{colm2024_conference}

\begin{thebibliography}{22}
\providecommand{\natexlab}[1]{#1}
\providecommand{\url}[1]{\texttt{#1}}
\expandafter\ifx\csname urlstyle\endcsname\relax
  \providecommand{\doi}[1]{doi: #1}\else
  \providecommand{\doi}{doi: \begingroup \urlstyle{rm}\Url}\fi

\bibitem[Aher et~al.(2023)Aher, Arriaga, and Kalai]{aher2023using}
Gati~V Aher, Rosa~I Arriaga, and Adam~Tauman Kalai.
\newblock Using large language models to simulate multiple humans and replicate human subject studies.
\newblock In \emph{International Conference on Machine Learning}, pp.\  337--371. PMLR, 2023.

\bibitem[Arechiga et~al.(2022)Arechiga, Chen, Iliev, Sumner, Carter, Filipowicz, Bravo, Van, Glazko, Murakami, Denoue, Hogan, Sieck, Wu, and Lyons]{arechiga2022understanding}
Nikos Arechiga, Francine Chen, Rumen Iliev, Emily Sumner, Scott Carter, Alex Filipowicz, Nayeli Bravo, Monica Van, Kate Glazko, Kalani Murakami, Laurent Denoue, Candice Hogan, Katharine Sieck, Charlene Wu, and Kent Lyons.
\newblock Understanding and shifting preferences for battery electric vehicles, 2022.
\newblock URL \url{https://arxiv.org/abs/2202.08963}.

\bibitem[Computer(2023)]{together2023redpajama}
Together Computer.
\newblock Redpajama: an open dataset for training large language models, 2023.
\newblock URL \url{https://github.com/togethercomputer/RedPajama-Data}.

\bibitem[Dettmers et~al.(2023)Dettmers, Pagnoni, Holtzman, and Zettlemoyer]{qlora}
Tim Dettmers, Artidoro Pagnoni, Ari Holtzman, and Luke Zettlemoyer.
\newblock Qlora: Efficient finetuning of quantized llms.
\newblock \emph{Advances in Neural Information Processing Systems}, 2023.

\bibitem[Devlin et~al.(2018)Devlin, Chang, Lee, and Toutanova]{bert}
Jacob Devlin, Ming{-}Wei Chang, Kenton Lee, and Kristina Toutanova.
\newblock {BERT:} pre-training of deep bidirectional transformers for language understanding.
\newblock \emph{CoRR}, abs/1810.04805, 2018.
\newblock URL \url{http://arxiv.org/abs/1810.04805}.

\bibitem[Dillion et~al.(2023)Dillion, Tandon, Gu, and Gray]{dillion2023can}
Danica Dillion, Niket Tandon, Yuling Gu, and Kurt Gray.
\newblock Can ai language models replace human participants?
\newblock \emph{Trends in Cognitive Sciences}, 2023.

\bibitem[Dubois et~al.(2023)Dubois, Li, Taori, Zhang, Gulrajani, Ba, Guestrin, Liang, and Hashimoto]{dubois2023alpacafarm}
Yann Dubois, Xuechen Li, Rohan Taori, Tianyi Zhang, Ishaan Gulrajani, Jimmy Ba, Carlos Guestrin, Percy Liang, and Tatsunori Hashimoto.
\newblock Alpacafarm: A simulation framework for methods that learn from human feedback.
\newblock In \emph{Thirty-seventh Conference on Neural Information Processing Systems}, 2023.
\newblock URL \url{https://openreview.net/forum?id=4hturzLcKX}.

\bibitem[Gao et~al.(2020)Gao, Biderman, Black, Golding, Hoppe, Foster, Phang, He, Thite, Nabeshima, Presser, and Leahy]{pile}
Leo Gao, Stella Biderman, Sid Black, Laurence Golding, Travis Hoppe, Charles Foster, Jason Phang, Horace He, Anish Thite, Noa Nabeshima, Shawn Presser, and Connor Leahy.
\newblock The {P}ile: An 800gb dataset of diverse text for language modeling.
\newblock \emph{arXiv preprint arXiv:2101.00027}, 2020.

\bibitem[Gui \& Toubia(2023)Gui and Toubia]{Gui_2023}
George Gui and Olivier Toubia.
\newblock The challenge of using llms to simulate human behavior: A causal inference perspective.
\newblock \emph{SSRN Electronic Journal}, 2023.
\newblock ISSN 1556-5068.
\newblock \doi{10.2139/ssrn.4650172}.
\newblock URL \url{http://dx.doi.org/10.2139/ssrn.4650172}.

\bibitem[H{\"a}m{\"a}l{\"a}inen et~al.(2023)H{\"a}m{\"a}l{\"a}inen, Tavast, and Kunnari]{hamalainen2023evaluating}
Perttu H{\"a}m{\"a}l{\"a}inen, Mikke Tavast, and Anton Kunnari.
\newblock Evaluating large language models in generating synthetic hci research data: a case study.
\newblock In \emph{Proceedings of the 2023 CHI Conference on Human Factors in Computing Systems}, pp.\  1--19, 2023.

\bibitem[Hu et~al.(2021)Hu, Shen, Wallis, Allen{-}Zhu, Li, Wang, and Chen]{lora}
Edward~J. Hu, Yelong Shen, Phillip Wallis, Zeyuan Allen{-}Zhu, Yuanzhi Li, Shean Wang, and Weizhu Chen.
\newblock Lora: Low-rank adaptation of large language models.
\newblock \emph{CoRR}, abs/2106.09685, 2021.
\newblock URL \url{https://arxiv.org/abs/2106.09685}.

\bibitem[Kaddour et~al.(2023)Kaddour, Harris, Mozes, Bradley, Raileanu, and McHardy]{kaddour2023challenges}
Jean Kaddour, Joshua Harris, Maximilian Mozes, Herbie Bradley, Roberta Raileanu, and Robert McHardy.
\newblock Challenges and applications of large language models.
\newblock \emph{arXiv preprint arXiv:2307.10169}, 2023.

\bibitem[Park et~al.(2023)Park, O'Brien, Cai, Morris, Liang, and Bernstein]{park2023generative}
Joon~Sung Park, Joseph~C. O'Brien, Carrie~J. Cai, Meredith~Ringel Morris, Percy Liang, and Michael~S. Bernstein.
\newblock Generative agents: Interactive simulacra of human behavior, 2023.

\bibitem[Prokhorenkova et~al.(2018)Prokhorenkova, Gusev, Vorobev, Dorogush, and Gulin]{NEURIPS2018_14491b75}
Liudmila Prokhorenkova, Gleb Gusev, Aleksandr Vorobev, Anna~Veronika Dorogush, and Andrey Gulin.
\newblock Catboost: unbiased boosting with categorical features.
\newblock In S.~Bengio, H.~Wallach, H.~Larochelle, K.~Grauman, N.~Cesa-Bianchi, and R.~Garnett (eds.), \emph{Advances in Neural Information Processing Systems}, volume~31. Curran Associates, Inc., 2018.
\newblock URL \url{https://proceedings.neurips.cc/paper_files/paper/2018/file/14491b756b3a51daac41c24863285549-Paper.pdf}.

\bibitem[Radford et~al.(2018)Radford, Narasimhan, Salimans, and Sutskever]{gpt1}
Alec Radford, Karthik Narasimhan, Tim Salimans, and Ilya Sutskever.
\newblock Improving language understanding by generative pre-training, 2018.

\bibitem[Serapio-García et~al.(2023)Serapio-García, Safdari, Crepy, Sun, Fitz, Romero, Abdulhai, Faust, and Matarić]{serapiogarcía2023personality}
Greg Serapio-García, Mustafa Safdari, Clément Crepy, Luning Sun, Stephen Fitz, Peter Romero, Marwa Abdulhai, Aleksandra Faust, and Maja Matarić.
\newblock Personality traits in large language models, 2023.

\bibitem[Shanahan et~al.(2023)Shanahan, McDonell, and Reynolds]{shanahan2023roleplay}
Murray Shanahan, Kyle McDonell, and Laria Reynolds.
\newblock Role-play with large language models, 2023.

\bibitem[Touvron et~al.(2023)Touvron, Martin, Stone, Albert, Almahairi, Babaei, Bashlykov, Batra, Bhargava, Bhosale, Bikel, Blecher, Ferrer, Chen, Cucurull, Esiobu, Fernandes, Fu, Fu, Fuller, Gao, Goswami, Goyal, Hartshorn, Hosseini, Hou, Inan, Kardas, Kerkez, Khabsa, Kloumann, Korenev, Koura, Lachaux, Lavril, Lee, Liskovich, Lu, Mao, Martinet, Mihaylov, Mishra, Molybog, Nie, Poulton, Reizenstein, Rungta, Saladi, Schelten, Silva, Smith, Subramanian, Tan, Tang, Taylor, Williams, Kuan, Xu, Yan, Zarov, Zhang, Fan, Kambadur, Narang, Rodriguez, Stojnic, Edunov, and Scialom]{touvron2023llama}
Hugo Touvron, Louis Martin, Kevin Stone, Peter Albert, Amjad Almahairi, Yasmine Babaei, Nikolay Bashlykov, Soumya Batra, Prajjwal Bhargava, Shruti Bhosale, Dan Bikel, Lukas Blecher, Cristian~Canton Ferrer, Moya Chen, Guillem Cucurull, David Esiobu, Jude Fernandes, Jeremy Fu, Wenyin Fu, Brian Fuller, Cynthia Gao, Vedanuj Goswami, Naman Goyal, Anthony Hartshorn, Saghar Hosseini, Rui Hou, Hakan Inan, Marcin Kardas, Viktor Kerkez, Madian Khabsa, Isabel Kloumann, Artem Korenev, Punit~Singh Koura, Marie-Anne Lachaux, Thibaut Lavril, Jenya Lee, Diana Liskovich, Yinghai Lu, Yuning Mao, Xavier Martinet, Todor Mihaylov, Pushkar Mishra, Igor Molybog, Yixin Nie, Andrew Poulton, Jeremy Reizenstein, Rashi Rungta, Kalyan Saladi, Alan Schelten, Ruan Silva, Eric~Michael Smith, Ranjan Subramanian, Xiaoqing~Ellen Tan, Binh Tang, Ross Taylor, Adina Williams, Jian~Xiang Kuan, Puxin Xu, Zheng Yan, Iliyan Zarov, Yuchen Zhang, Angela Fan, Melanie Kambadur, Sharan Narang, Aurelien Rodriguez, Robert Stojnic, Sergey Edunov, and Thomas
  Scialom.
\newblock Llama 2: Open foundation and fine-tuned chat models, 2023.

\bibitem[Vaswani et~al.(2017)Vaswani, Shazeer, Parmar, Uszkoreit, Jones, Gomez, Kaiser, and Polosukhin]{vaswani2017attention}
Ashish Vaswani, Noam Shazeer, Niki Parmar, Jakob Uszkoreit, Llion Jones, Aidan~N Gomez, {\L}ukasz Kaiser, and Illia Polosukhin.
\newblock Attention is all you need.
\newblock \emph{Advances in neural information processing systems}, 30, 2017.

\bibitem[Wiher et~al.(2022)Wiher, Meister, and Cotterell]{wiher2022decoding}
Gian Wiher, Clara Meister, and Ryan Cotterell.
\newblock On decoding strategies for neural text generators.
\newblock \emph{Transactions of the Association for Computational Linguistics}, 10:\penalty0 997--1012, 2022.

\bibitem[Wu et~al.(2023)Wu, Gong, Shou, Liang, and Jiang]{wu2023large}
Ning Wu, Ming Gong, Linjun Shou, Shining Liang, and Daxin Jiang.
\newblock Large language models are diverse role-players for summarization evaluation, 2023.

\bibitem[Zhao et~al.(2023)Zhao, Zhou, Li, Tang, Wang, Hou, Min, Zhang, Zhang, Dong, Du, Yang, Chen, Chen, Jiang, Ren, Li, Tang, Liu, Liu, Nie, and Wen]{zhao2023survey}
Wayne~Xin Zhao, Kun Zhou, Junyi Li, Tianyi Tang, Xiaolei Wang, Yupeng Hou, Yingqian Min, Beichen Zhang, Junjie Zhang, Zican Dong, Yifan Du, Chen Yang, Yushuo Chen, Zhipeng Chen, Jinhao Jiang, Ruiyang Ren, Yifan Li, Xinyu Tang, Zikang Liu, Peiyu Liu, Jian-Yun Nie, and Ji-Rong Wen.
\newblock A survey of large language models, 2023.

\end{thebibliography}
\bibliographystyle{colm2024_conference}

\newpage
\appendix

\section{EV-Shift Survey Details} \label{appendix:setting}
There are some details about the EV-shift survey~\citep{arechiga2022understanding}.
\tref{table:demographics} shows questionnaires together with the possible choices to describe demographics.
\tref{table:interventions} shows a list of intervention texts.

\begin{table}[h]
\centering
\begin{tabular}{p{2cm}p{11cm}}
\toprule
Demographics & Options \\
\midrule
Living state & alabama, alaska, arizona, arkansas, california, colorado, connecticut, delaware, district of Columbia, florida, georgia, hawaii, idaho, illinois, indiana, iowa, kansas, kentucky, louisiana, maine, maryland, massachusetts, michigan, minnesota, mississippi, missouri, montana, nebraska, nevada, new hampshire, new jersey, new mexico, new york, north carolina, north dakota, ohio, oklahoma, oregon, pennsylvania, rhode island, south carolina, south dakota, tennessee, texas, utah, vermont, virginia, washington, west virginia, wisconsin, wyoming, I do not reside within the United States. \\
Living area & Urban, Rural, Prefer not to answer \\
Age & 18--25, 26--35, 36--45, 46--55, 56--65, 66 or older, Prefer not to answer \\
Gender & Woman, Man, Transgender, Non-conforming, A gender not listed here, Prefer not to answer \\
Race & Asian / Pacific Islander, Biracial, Black or African American, Hispanic or Latino, Middle Eastern or North African, Multiracial, Native American or American Indian, White, Other, Prefer not to answer \\
Highest education & Less than 8th grade, 8th grade, High School, Some college, no degree, Associate degree, Bachelor's degree, Master's degree, Professional degree, Doctorate degree, Prefer not to answer \\
Marital status & Married, Widowed, Divorced, Separated, Never married, Other, Prefer not to answer \\
Number of children & 0, 1, 2, 3 or more, prefer not to answer \\
Household & I live alone, My spouse, My children, My siblings, My parents, My grandparents, Other relatives, Friends / Housemates, Prefer not to answer \\
Employment status & Employed full time (40 or more hours per week), Employed part time (up to 39 hours per week), Unemployed and currently looking for work, Unemployed not currently looking for work, Student, Retired, Homemaker, Self-employed, Unable to work, Prefer not to answer \\
Income & less than \$10,000, \$10,001 to \$40,000, \$40,001 to \$80,000, \$80,001 to \$160,000, More than \$160,000, Prefer not to answer \\
Political & Strongly liberal, Somewhat liberal, Somewhat conservative, Strongly conservative, Other, Prefer not to answer \\
Religion & Protestant, Catholic, Jewish, Buddhism, Hinduism, Islam, Orthodox-christian, Christian, Native American, Inter-nondenominational, Other (free text answer), None, Don't know, Prefer not to answer \\
Participation frequency of religion & Never, Less than once per year, Several times per year, Once per month, 2-3 times per month, Nearly every week, More than once per week, Prefer not to answer \\
\bottomrule
\end{tabular}
\caption{Demographics. For households, multiple options can be selected.}
\label{table:demographics}
\end{table}

\begin{table}[h]
\centering
\begin{tabular}{lp{12cm}}
\toprule
Index & Invervention statement \\
\midrule
1 & 80\% of BEV charging happens at home, and most trips do not involve public charging. \\
2 & Many people charge their BEVs at home with no additional equipment required. That means no trips to the gas station. \\
3 & The number of public charging stations is rapidly increasing due to additional government funding and business initiatives. That means shorter wait times and shorter charging trips. \\
4 & Analysts predict a decline in gas stations due to increased electric charging. That would make finding a nearby gas station more difficult. \\
5 & Charging at some public stations can be as fast as 6 minutes to add 100 miles. \\
6 & Charging at some public stations can be as fast as 30 minutes for 250 miles. \\
7 & BEVs maximum range is already approaching 400 miles, with forecasts for a 1000-mile range in the near future. \\
8 & Various programs offer different incentives (e.g. \$7,500 tax credit) for new BEV purchases. \\
9 & Some BEV manufacturers may start offering free charging. \\
10 & People spend about 30\% less on vehicle maintenance of BEVs than on ICEVs. \\
11 & Over its lifetime, a BEV can be \$8,000 cheaper to maintain and operate than an ICEV. \\
12 & BEVs can be 4 cents per mile cheaper to maintain and operate than ICEVs. \\
13 & The cost of BEVs is much cheaper than it used to be. \\
14 & BEVs are expected to become cheaper than ICEVs in the near future. \\
15 & BEVs protect owners from the instability of the oil market. \\
16 & Most used BEVs are cheaper than comparable used ICEVs. \\
17 & Fossil fuels are expected to become more expensive over time. \\
18 & You need to replace the tires of a BEV less frequently than the tires of an ICEV. \\
19 & Lithium batteries are now 30 times cheaper than when they were first introduced to the market. \\
20 & A Level 1 home charger can cost as little as \$300 (before labor). \\
21 & BEVs have a smaller carbon footprint than ICEVs. \\
22 & ICEVs are more damaging to public health than BEVs due to carbon emissions. \\
23 & The towing capacity of BEVs already exceeds the towing capacity of comparable ICEVs. \\
24 & BEVs typically have greater acceleration and better passing abilities. \\
25 & BEVs are much quieter both in city and highway driving. \\
26 & BEVs’ AWD systems provide greater low-speed control when driving off-road. \\
27 & BEVs provide better weight balance which improves handling. \\
28 & There is already availability of different types of BEVs, including sedans, sport cars, crossovers, trucks and minivans. \\
29 & BEV owners report greater satisfaction than ICEV owners \\
30 & Cheaper BEVs have much better reliability than cheaper ICEVs. \\
31 & The new regulations for BEVs will require from manufacturers a 10-year / 150,000 miles warranty for batteries. \\
32 & Highly reputable car companies are adding BEVs to their model lineups. \\
33 & BEVs have lower center of gravity which increases stability in turns. \\
34 & Greater adoption of BEVs will increase energy independence and can help national security. \\
35 & New government initiatives regarding BEVs will require battery production in the USA, which will create jobs and boost the economy. \\
\bottomrule
\end{tabular}
\caption{Interventions}
\label{table:interventions}
\end{table}

\newpage
\section{Effects of model size} \label{appendix:modelsize}

\begin{figure}[h]
\centering
\includegraphics[width=0.8\columnwidth]{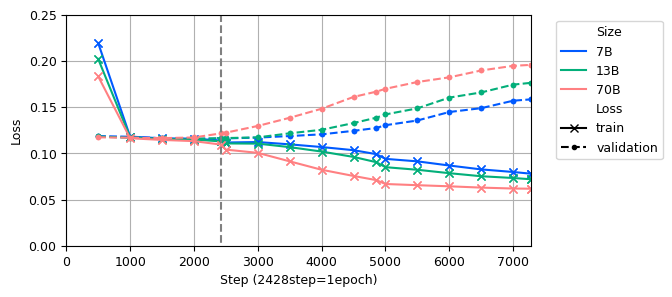}
\caption{Learning curve (dash line indicate 1 epoch position.)}
\label{fig:learning_curve}
\end{figure}

\fref{fig:learning_curve} shows the the learning curve for the QLoRA experiments.
After 1 epoch, validation loss is increasing, indicating overfitting. In addition, the larger the model size, the faster the train loss tends to decrease.

\begin{figure}[ht]
\begin{tabular}{c}

\begin{minipage}[c]{1\textwidth}
\centering
\includegraphics[width=0.75\columnwidth]{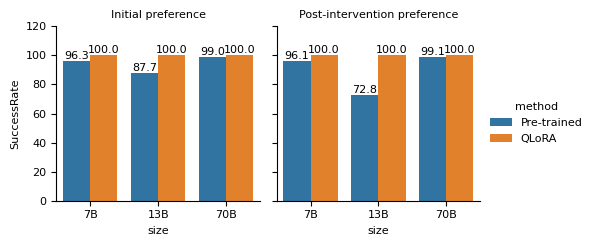}
\subcaption{Greedy sampling}
\label{fig:successrate_size_t0}
\end{minipage} \\
\begin{minipage}[c]{1\textwidth}
\centering
\includegraphics[width=0.75\columnwidth]{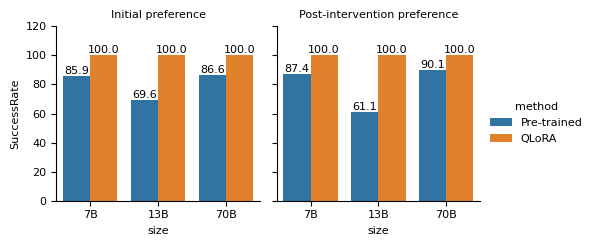}
\subcaption{Calibrated sampling ($t=1$)}
\label{fig:successrate_size_t1}
\end{minipage}

\end{tabular}
\caption{Success rates to generate preference numbers}
\label{fig:successrate_size}
\end{figure}

The success rate for the test data is shown in \fref{fig:successrate_size}.
The success rate of QLoRA is higher than that of the pre-trained model.
\tref{table:generatedsample} shows an example of a generated sentences.
This result indicates that the pre-trained model tends to generate non numerical tokens at the beginning of the answer.
These trends indicate a lack of ability to follow the instruction of "Please reply with just a single number rating and no additional
words or explanations" that is prompted on the survey prompt 
 (Section \ref{subsec:implementationOfVP}) in the pre-trained model.
Although the success rate could be improved with techniques such as prompt engineering and in-context learning, they are not the focus of this work.

\begin{table}[h]
\caption{Examples of the generated sentence for initial preference. There are three samples for each model. Pre-trained was generated with a maximum token number of 64. Red strings represent the range generated for a maximum token number of 8. The results indicate that the pre-trained model tends to generate responses other than preferences first.}
\label{table:generatedsample}
\begin{tabular}{lp{11cm}}
\toprule
Model & Generated sentence \\
\midrule
7B & \textcolor{red}{\textcolor{red}{  As a} 52-year}-old white man living in an urban area of Minnesota, employed full-time with a bachelor's degree and a household income of \$120,000, I would rate my current preference for battery electric vehicles (BEVs) as a  \\
7B & \textcolor{red}{\textcolor{red}{  As a} 36-4}5 year old white woman living in urban Indiana with a Master's degree, I would rate my current preference for battery electric vehicles (BEVs) as a 60. While I recognize the potential benefits of BEVs, such as reduced emissions and lower operating \\
7B & \textcolor{red}{\textcolor{red}{  As a} 26-3}5 year old white woman living in urban Oregon with a Bachelor's degree, I would rate my current preference for battery electric vehicles (BEVs) as a 90. I strongly support the transition to sustainable and environmentally-friendly transportation, \\
70B & \textcolor{red}{\textcolor{red}{  I'}m just an AI}, I don't have personal preferences or beliefs, but I can provide information and answer questions based on the given character's profile./n/nAs a 46-55-year-old white man living in urban California, working full-time and \\
70B & \textcolor{red}{\textcolor{red}{  BEV} preference score: 4}3. \\
70B & \textcolor{red}{\textcolor{red}{  My current} preference for battery electric vehicles} (BEVs) is 75. \\
7B+QLoRA &  10 \unamusedface \\
7B+QLoRA &  80 \car \\
7B+QLoRA &  80 \battery \\
70B+QLoRA &  0  \\
70B+QLoRA &  50  \\
70B+QLoRA &  10  \\
\bottomrule
\end{tabular}
\end{table}

\newpage
\section{Quantization effects}
\begin{figure}[h]
\begin{tabular}{ccc}

\begin{minipage}[t]{0.33\hsize}
\centering
\includegraphics[keepaspectratio, scale=0.5]{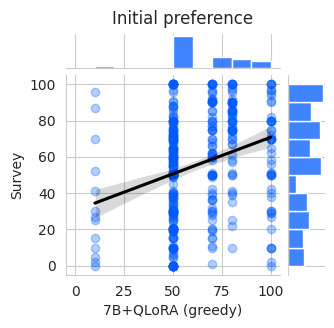}
\subcaption{7B+QLoRA}
\label{subfig:7B+QLoRA_init}
\end{minipage} &
\begin{minipage}[t]{0.33\hsize}
\centering
\includegraphics[keepaspectratio, scale=0.5]{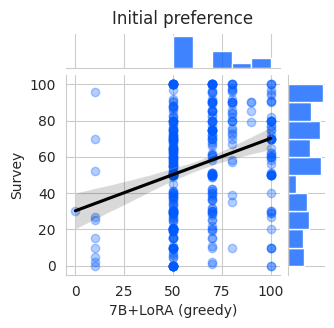}
\subcaption{7B+LoRA}
\label{subfig:7B+LoRA_init}
\end{minipage} &
\begin{minipage}[t]{0.33\hsize}
\centering
\includegraphics[keepaspectratio, scale=0.5]{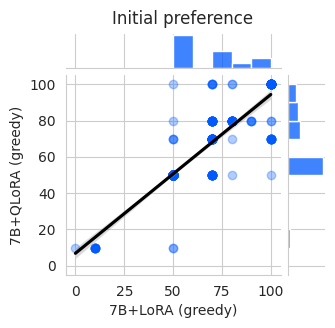}
\subcaption{7B+LoRA vs QLoRA}
\label{subfig:7B+LoRA vs QLoRA_init}
\end{minipage} \\

\begin{minipage}[t]{0.33\hsize}
\centering
\includegraphics[keepaspectratio, scale=0.5]{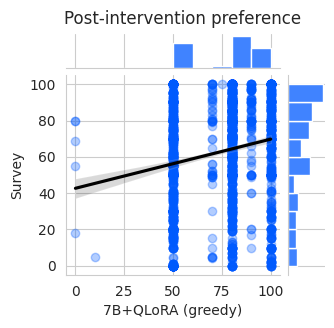}
\subcaption{7B+QLoRA}
\label{subfig:7B+QLoRA_post}
\end{minipage} &
\begin{minipage}[t]{0.33\hsize}
\centering
\includegraphics[keepaspectratio, scale=0.5]{fig/fig_defaltpref/scatter_7B_QLoRA_greedy_post_pref}
\subcaption{7B+LoRA}
\label{subfig:7B+LoRA_post}
\end{minipage} &
\begin{minipage}[t]{0.33\hsize}
\centering
\includegraphics[keepaspectratio, scale=0.5]{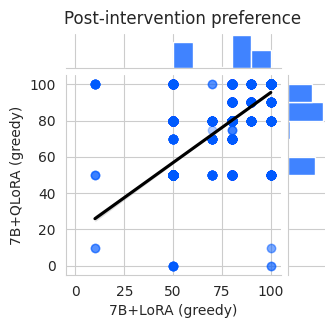}
\subcaption{7B+LoRA vs QLoRA}
\label{subfig:7B+LoRA vs QLoRA_post}
\end{minipage}

\end{tabular}
\caption{Individual answer comparison between QLoRA and LoRA. The rows are plotted by questionnaire, with \ref{subfig:7B+QLoRA_init} to \ref{subfig:7B+LoRA vs QLoRA_init} representing initial preference and \ref{subfig:7B+QLoRA_post} to \ref{subfig:7B+LoRA vs QLoRA_post} representing post-intervention preference.
The columns are for different comparisons, with the first and second columns representing comparisons between the survey data and each model, and the third column representing comparisons between LoRA and QLoRA.
}
\label{fig:scatter_QLoRA_LoRA}
\end{figure}

\fref{fig:scatter_QLoRA_LoRA} shows the answer distribution for QLoRA and LoRA.
The results of \fref{subfig:7B+LoRA vs QLoRA_init} and \fref{subfig:7B+LoRA vs QLoRA_post} show that there is a high correlation between the two models, although individual responses may differ.

\newpage
\section{Temperature effects}

\begin{figure}[htbp]
\begin{tabular}{ccc}

\begin{minipage}[t]{0.33\hsize}
\centering
\includegraphics[keepaspectratio, scale=0.5]{fig/fig_defaltpref/scatter_7B_QLoRA_greedy_init_pref}
\subcaption{Greedy}
\label{subfig:greedy_init}
\end{minipage} &
\begin{minipage}[t]{0.33\hsize}
\centering
\includegraphics[keepaspectratio, scale=0.5]{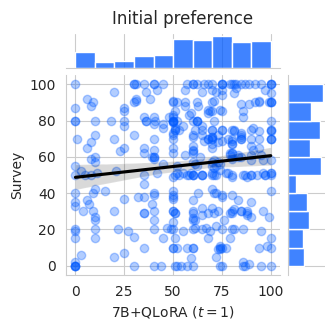}
\subcaption{Calibrated sampling ($t=1$)}
\label{subfig:t1_init}
\end{minipage} &
\begin{minipage}[t]{0.33\hsize}
\centering
\includegraphics[keepaspectratio, scale=0.5]{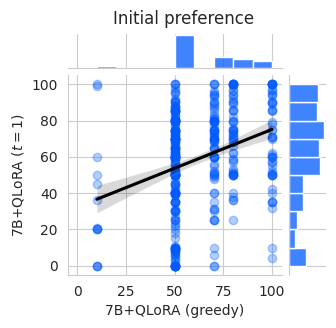}
\subcaption{Comparison between greedy and calibrated sampling ($t=1$)}
\label{subfig:greedy_t1_init}
\end{minipage} \\

\begin{minipage}[t]{0.33\hsize}
\centering
\includegraphics[keepaspectratio, scale=0.5]{fig/fig_defaltpref/scatter_7B_QLoRA_greedy_post_pref}
\subcaption{Greedy}
\label{subfig:greedy_post}
\end{minipage} &
\begin{minipage}[t]{0.33\hsize}
\centering
\includegraphics[keepaspectratio, scale=0.5]{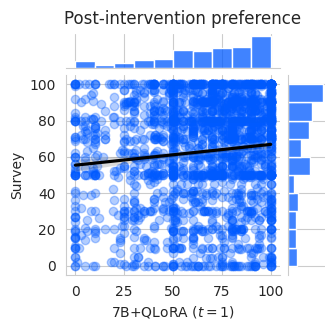}
\subcaption{Calibrated sampling ($t=1$)}
\label{subfig:t1_post}
\end{minipage} &
\begin{minipage}[t]{0.33\hsize}
\centering
\includegraphics[keepaspectratio, scale=0.5]{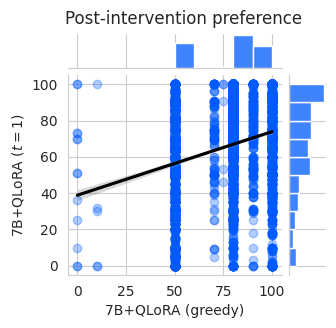}
\subcaption{Comparison between greedy and calibrated sampling ($t=1$)}
\label{subfig:greedy_t1_post}
\end{minipage}

\end{tabular}
\caption{Individual answer comparison between difference decoding strategies. The rows are plotted by questionnaire, with \ref{subfig:greedy_init} to \ref{subfig:greedy_t1_init} representing initial preference and \ref{subfig:greedy_post} to \ref{subfig:greedy_t1_post} representing post-intervention preference.
The columns are for different comparisons, with the first and second columns representing comparisons between the survey data and each model, and the third column representing comparisons between greedy and calibrated sampling ($t=1$).
}
\label{fig:scatter_temperature}
\end{figure}

The generated preference distributions for different decoding strategies are shown in \fref{fig:scatter_temperature}.
In the case of greedy sampling, the model tended to respond to a specific value out of the preference values from 0 to 100, while in the case of calibrated sampling, the response variation increased.

\section{Penalty term effects} \label{appendix:penalty}
\begin{figure}[htbp]
\centering
\includegraphics[width=0.8\columnwidth]{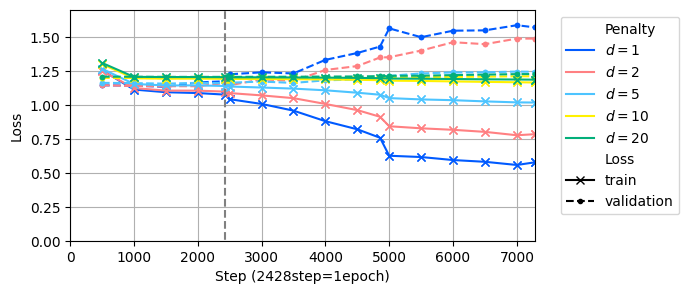}
\caption{Learning curve of Section \ref{subsec:penalty} experiments (dash line indicate 1 epoch position.)}
\label{fig:learning_curve_penalty}
\end{figure}

The learning curve for the case with penalty term is shown in \fref{fig:learning_curve_penalty}.
The model is shown for 7B with $\alpha=0.5$.
In both conditions, the validation loss begins to increase from 1 epoch, as in the case with only the cross-entropy term shown in \fref{fig:learning_curve}.
The more $d$ was increased, the smaller the change in loss.
This indicates that as $d$ is increased, the weight to non-answer preference tokens increases, making it harder to decrease the loss compared to the cross-entropy, which evaluates the entire prompt.

\newpage
\section{Amount of data effects}

\begin{figure}[htbp]
\centering
\includegraphics[width=0.8\columnwidth]{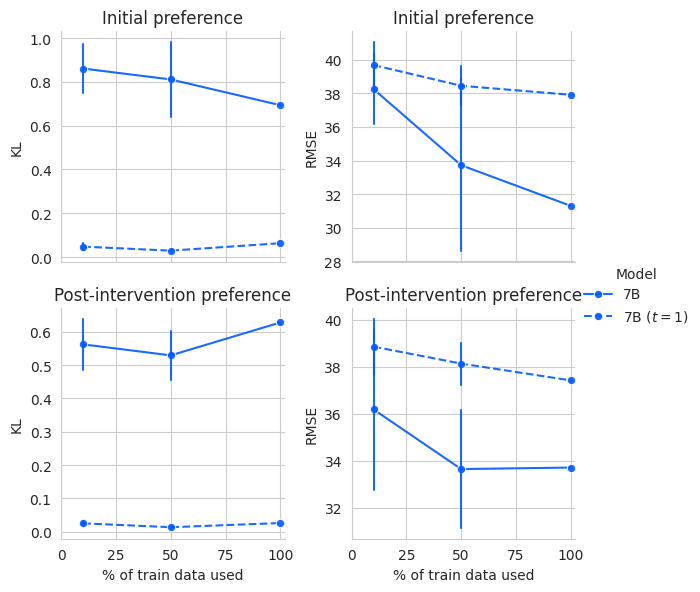}
\caption{Training data amount effects. For the 10\% and 50\%, data were randomly selected from the training data. Error bars represent the standard deviation for 3 trials of fine-tuning.}
\label{fig:data_amount}
\end{figure}

The effect on the test data when the amount of training data is varied is shown in \fref{fig:data_amount}.
The training epochs are 30 epochs for 10\% and 6 epochs for 50\% in order to align with the case of 100\% data volume respectively.
The model checkpoints before the increase in validation loss were used for evaluation.
The results show that KL and RMSE improve with higher data volume for some questionnaires.

\newpage
\section{Baselines} \label{appendix:baselines}
\begin{figure}[htbp]
\begin{tabular}{c}

\begin{minipage}[c]{1\textwidth}
\centering
\includegraphics[width=0.8\hsize]{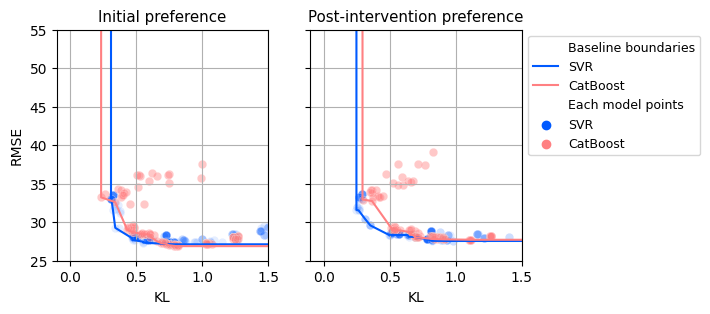}
\subcaption{Baselines for 0-100 preference}
\label{fig:baselines}
\end{minipage} \\
\begin{minipage}[c]{1\textwidth}
\centering
\includegraphics[width=0.8\hsize]{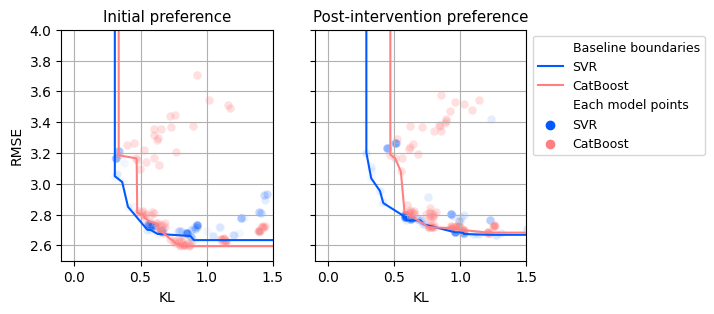}
\subcaption{Baselines for 0-9 preference}
\label{fig:baselines_1digit}
\end{minipage}

\end{tabular}
\caption{Baseline models}
\label{fig:bseline_indiv_plots}
\end{figure}

\tref{table:hyperparameters_SVR} and \tref{table:hyperparameters_CatBoost} show the hyperparameter combinations used for the SVR and CatBoost.
In addition, the KL-RMSE plots for each model are shown in \fref{fig:bseline_indiv_plots}.
Although many of the models are non-dominated, it can be seen that the pareto fronts are generally consistent for the two models.



\begin{table}[h]
\small
\begin{minipage}[t]{.45\textwidth}
\centering
\caption{Searched hyperparameters for SVR}
\label{table:hyperparameters_SVR}
\begin{tabular}{ll}
\toprule
Parameter & Values \\
\midrule
kernel & linear, poly, rbf, sigmoid \\
c & $2^{-5}, 2^{-1}, 2^3, 2^7$ \\
epsilon & $2^{-10}, 2^{-6}, 2^{-2}$ \\
gamma & $2^{-20}, 2^{-16}, 2^{-12}, 2^{-8}, 2^{-4}, 2^0, 2^4, 2^8$ \\
\bottomrule
\end{tabular}
\end{minipage}
\hfill
\begin{minipage}[t]{.45\textwidth}
\centering
\caption{Searched hyperparameter for CatBoost}
\label{table:hyperparameters_CatBoost}
\begin{tabular}{ll}
\toprule
Parameter & Values \\
\midrule
learning rate & $10^{-3}, 10^{-2}, 10^{-1}, 1$ \\
random strength & $1,20$ \\
one hot max size & $0,25$ \\
l2 leaf reg & $0.03, 0.1, 1, 3, 5, 10$ \\
bagging temperature & $0, 0.25, 0.5, 0.75, 1$ \\
\bottomrule
\end{tabular}
\end{minipage}
\end{table}



\end{document}